\documentclass[10pt,journal,compsoc]{IEEEtran}
\ifCLASSOPTIONcompsoc
  \usepackage[nocompress]{cite}
\else
  \usepackage{cite}
\fi
\ifCLASSINFOpdf
\usepackage[pdftex]{graphicx}
\else
\fi
\usepackage{amsmath}
\usepackage{mdwmath}
\usepackage{mdwtab}
\usepackage{eqparbox}
\usepackage{algorithmic}

\usepackage{array}
\newcommand\MYhyperrefoptions{bookmarks=true,bookmarksnumbered=true,
pdfpagemode={UseOutlines},plainpages=false,pdfpagelabels=true,
colorlinks=true,linkcolor={black},citecolor={black},urlcolor={black},
pdftitle={ARBEx: Attentive Feature Extraction with Reliability Balancing for Robust Facial Expression Learning},
pdfsubject={Typesetting},
pdfauthor={Michael D. Shell},
pdfkeywords={Facial expression learning, Reliability balancing, Bias and uncertainty, Multi-head attention}}
\ifCLASSINFOpdf
\usepackage[\MYhyperrefoptions,pdftex]{hyperref}
\else
\usepackage[\MYhyperrefoptions,breaklinks=true,dvips]{hyperref}
\usepackage{breakurl}
\fi

\begin{document}
%
\title{ARBEx: Attentive Feature Extraction with Reliability Balancing for Robust Facial Expression Learning}
%
%
%
%

\author{Azmine Toushik Wasi$^*$,
        Karlo Šerbetar$^*$,
        Raima Islam$^*$,
        Taki Hasan Rafi$^*$,
        and Dong-Kyu Chae$^\dagger$
        
\IEEEcompsocitemizethanks{\IEEEcompsocthanksitem T. H. Rafi and D.K. Chae are with the Department Computer Science, Hanyang University, South Korea.\protect\\
E-mail:\{takihr, dongkyu\}@hanyang.ac.kr
\IEEEcompsocthanksitem A. T. Wasi is with Shahjalal University of Science and Technology, Bangladesh.\protect\\
E-mail: azminetoushik.wasi@gmail.com
\IEEEcompsocthanksitem S. Karlo was with University of Cambridge, UK.\protect\\
E-mail: serbetar.karlo@protonmail.com
\IEEEcompsocthanksitem R. Islam was with BRAC University, Bangladesh.\protect\\
E-mail: raima.islam@g.bracu.ac.bd
\IEEEcompsocthanksitem *These authors have contributed equally to this work.
\IEEEcompsocthanksitem $^\dagger$Corresponding Author.
\IEEEcompsocthanksitem Code: \href{https://github.com/takihasan/ARBEx}{\textcolor{blue}{https://github.com/takihasan/ARBEx}}.
\protect\\
}}

%
%

\markboth{\textbf{ARBEx}}%
{Shell \MakeLowercase{\textit{et al.}}: Bare Demo of IEEEtran.cls for Computer Society Journals}



\IEEEtitleabstractindextext{%
\begin{abstract}
In this paper, we introduce a framework ARBEx, a novel attentive feature extraction framework driven by Vision Transformer with reliability balancing to cope against poor class distributions, bias, and uncertainty in the facial expression learning (FEL) task. We reinforce several data pre-processing and refinement methods along with a window-based cross-attention ViT to squeeze the best of the data. We also employ learnable anchor points in the embedding space with label distributions and multi-head self-attention mechanism to optimize performance against weak predictions with reliability balancing, which is a strategy that leverages anchor points, attention scores, and confidence values to enhance the resilience of label predictions. To ensure correct label classification and improve the model’s discriminative power, we introduce anchor loss, which encourages large margins between anchor points. Additionally, the multi-head self-attention mechanism, which is also trainable, plays an integral role in identifying accurate labels. This approach provides critical elements for improving the reliability of predictions and has a substantial positive effect on final prediction capabilities. Our adaptive model can be integrated with any deep neural network to forestall challenges in various recognition tasks. Our strategy outperforms current state-of-the-art methodologies, according to extensive experiments conducted in a variety of contexts. 
\end{abstract}

}

\maketitle

\IEEEdisplaynontitleabstractindextext

%
\IEEEpeerreviewmaketitle

\IEEEraisesectionheading{\section{Introduction}\label{sec:introduction}}

\IEEEPARstart{O}{ne} of the most universal and significant methods that people communicate their emotions and intentions is through the medium of their facial expressions \cite{wang2020suppressing}. In recent years, facial expression learning (FEL) has garnered growing interest within the area of computer vision due to the fundamental importance of enabling computers to recognize interactions with humans and their emotional affect states. While FEL is a thriving and prominent research domain in human-computer interaction systems, its applications are also prevalent in healthcare, education, virtual reality, smart robotic systems, etc \cite{mao2023poster,ruan2021feature,she2021dive}. 

\begin{figure}[h!] 
\centering {\includegraphics[scale=0.35]{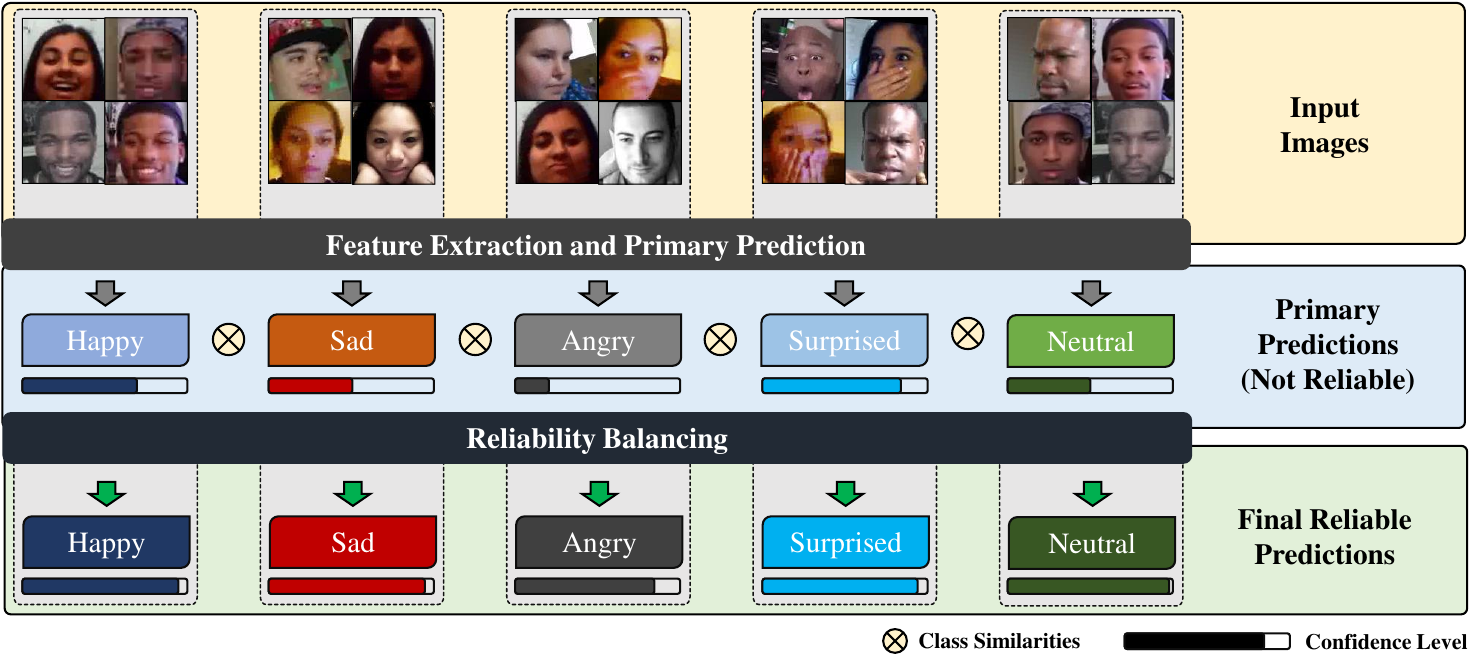}}
\caption{A synopsis of  \textbf{ARBEx}.  \textbf{Feature Extraction} provides feature maps to generate initial predictions. Confidence distributions of initial labels are mostly inconsistent, unstable and unreliable.  \textbf{Reliability Balancing} approach aids in stabilizing the distributions and addressing inconsistent and unreliable labeling.}\label{fig:2}
\end{figure}
The volume and quantity of large-scale benchmark FEL databases have significantly expanded in the past two decades \cite{du2014compound, li2017reliable, lucey2010extended, mollahosseini2017affectnet}, resulting in considerable improvement of recognition accuracy of some Convolutional Neural Network (CNN) methods, which integrated landmarks \cite{zhang2017facial}, prior knowledge \cite{chen2019facial, liu2013aware}, or image samples with optical flows  \cite{sun2019deep} for enhancement of the interpretability and performance \cite{chen2020label}. By separating the disturbances brought on by different elements, such as position, identity, lighting, and so on, several FEL approaches \cite{cai2018island,li2017reliable,ruan2020deep,xie2019deep} have also been developed to learn holistic expression aspects.

However, despite its recent outstanding performance, FEL is still considered as a difficult task due to a few reasons: (1) \textbf{ Global information.} Existing FEL methods fail to acknowledge global factors of input images due to the constraints of convolutional local receptive fields, (2)\textbf{ Inter-class similarity.} Several expression categories frequently include similar images with little differences between them, (3)\textbf{ Intra-class disparity.} Images from the same expression category might differ significantly from one another. For example, complexion, gender, image background, and an individual's age vary between instances, and (4)\textbf{ Sensitivity of scales.} Variations in image quality and resolution can often compromise the efficacy of deep learning networks when used without necessary precaution. Images from in-the-wild datasets and other FEL datasets come in a wide range of image sizes. Consequently, it is essential for FEL to provide consistent performance across scales \cite{vo2020pyramid}. 

In view of these difficulties and with the ascent of Transformers in the computer vision research field \cite{bi2021transformer}, numerous FEL techniques incorporated with Transformers have been developed which have achieved state-of-the-art (SOTA) results. 
Kim et al. \cite{kim2022facial} further developed Vision Transformer (ViT) to consolidate both global and local features so ViTs can be adjusted to FEL tasks. Furthermore, \cite{vo2020pyramid} addresses scale sensitivity. Transformer models with multi channel-based cross-attentive feature extraction such as POSTER \cite{zheng2022poster} and POSTER++ \cite{mao2023poster} tackle all of the aforementioned issues present in FEL, which use mutli-level cross attention based network powered by ViT to extract features. This architecture has surpassed many existing works in terms of performance owing to the help of its cross-fusion function, multi-scale feature extraction, and landmark-to-image branching method. The downsides of this complex design is that it easily overfits, does not employ heavy augmentation methods, nor does it deal with the certain bias, uncertainty awareness, and poor distribution of labels; problems commonly recognized in facial expression classification. 

To challenge this issue, we provide a novel reliability balancing approach where we place anchor points of different classes in the embeddings learned from \cite{mao2023poster}. These anchors have fixed labels and are used to calculate how similar the embeddings are to each label. We also add multi-head self-attention for the embeddings to find crucial components with designated weights to increase model reliability and robustness. This approach results in improved label distribution of the expressions along with stable confidence score for proper labeling in unreliable labels. To address the quick overfitting of enormous multi-level feature extraction procedures, we introduce heavy data augmentation and a robust training batch selection method to mitigate the risk of overfitting. 

In summary, our method offers unbiased and evenly distributed data to the image encoder, resulting in accurate feature maps. These feature maps are then utilized for drawing predictions with the assistance of reliability balancing section, ensuring robust outcomes regardless of potential bias and unbalance in the data and labels.
\subsection{Our Contributions}
Our contributions are summarized into four folds: 

\begin{itemize}
    \item We propose a novel approach \textbf{ARBEx}, a novel framework consisting multi-level attention-based feature extraction with reliability balancing for robust FEL with extensive data preprocessing and refinement methods to fight against biased data and poor class distributions.
    \item We propose \textbf{adaptive anchors in the embedding space and multi-head self-attention to increase reliability and robustness} of the model by correcting erroneous labels, providing more accurate and richer supervision for training the deep FEL network. We combine \textbf{relationships between anchors and weighted values from attention mechanism to stabilise class distributions} for poor predictions mitigating the issues of similarity in different classes effectively.
    \item Our streamlined data pipeline ensures \textbf{well-distributed quality input and output embeddings}, utilizing the full power of the Window-based Cross-Attention Vision Transformer providing robust feature maps to identify facial expressions in a confident manner. 
    \item Empirically, our \textbf{ARBEx} method is rigorously evaluated on diverse in-the-wild FEL databases. Experimental outcomes exhibit that our method consistently surpasses most of the state-of-art FEL systems. 
\end{itemize}

\begin{figure*}[t] 
\centering {\includegraphics[scale=0.5]{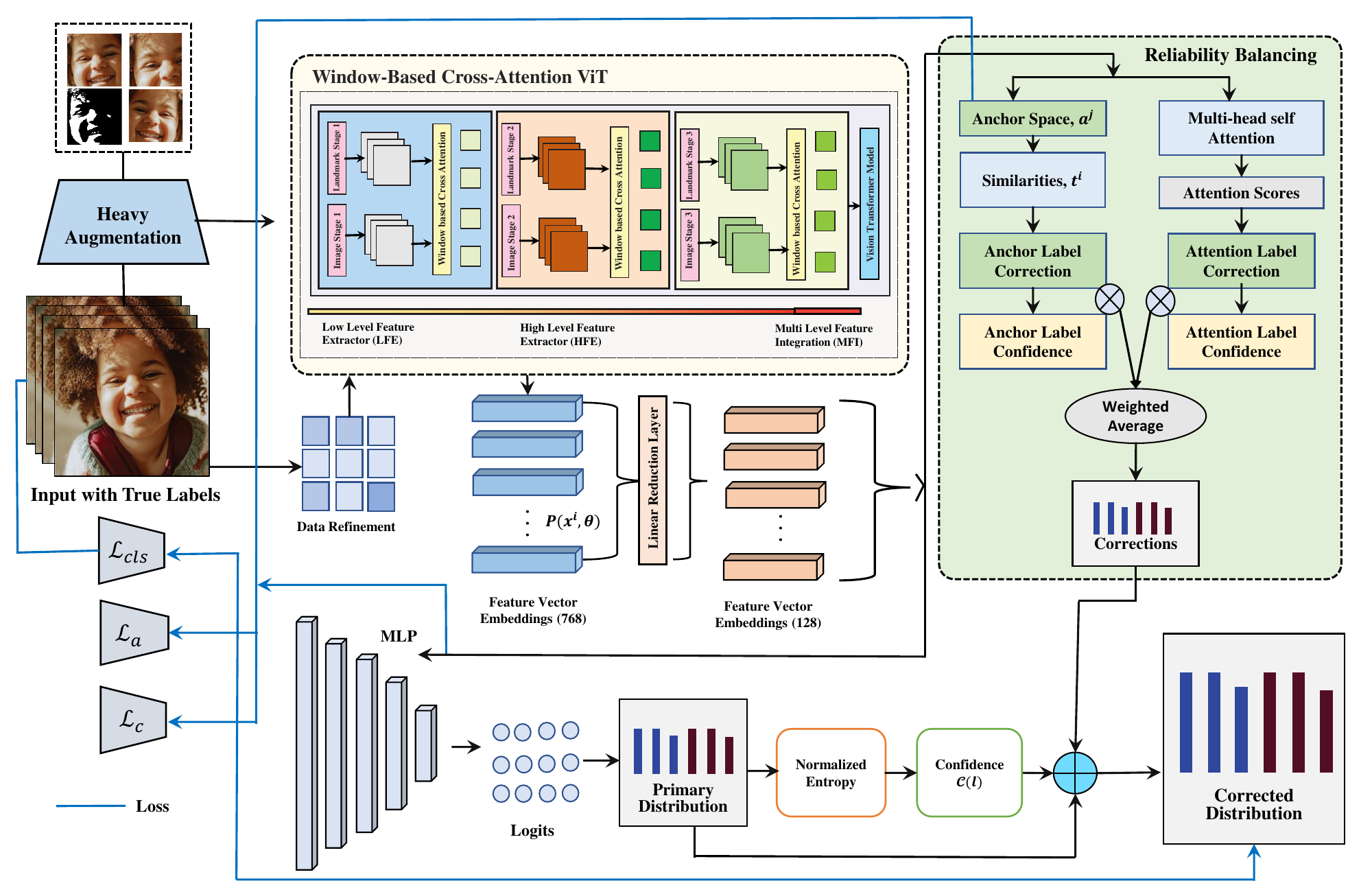}}
\caption{Pipeline of \textbf{ARBEx}. \textbf{Heavy Augmentation} is applied to the input images and \textbf{Data Refinement} method selects training batch with properly distributed classes for each epoch. \textbf{Window-Based Cross-Attention ViT} framework uses mutli-level feature extraction and integration to provide embeddings \textbf{(Feature Vectors).} \textbf{Linear Reduction Layer} reduces the feature vector size for fast modeling. \textbf{MLP} predicts the primary labels and \textbf{Confidence} is calculated from label distribution. \textbf{Reliability balancing} receives embeddings and processes in two ways. Firstly, it places anchors in the embedding space. It improves prediction probabilities by utilizing \textbf{trainable anchors} for searching similarities in embedding space. On the other way, \textbf{Multi-head self-attention} values are used to calculate label correction and confidence. \textbf{Weighted Average} of these two are used to calculate the final label \textbf{correction}. Using label correction, primary label distribution and confidence, final corrected label distribution is calculated, making the model more reliable.}\label{fig:2}
\end{figure*}

\section{Related Works}
In this section, we highlighted relevant works in FEL with respect to Transformers, uncertainty, and attention networks.
\subsection{Facial Expression Learning (FEL)} Most classic works related to FEL are used in a extensive variety, mainly in the computer vision and psychological science domains. In simple words, FEL is the task of labeling the expressions based on a facial image which consists of three phases, namely facial detection, feature extraction, and expression recognition \cite{wang2020suppressing}. In recent times, FEL systems have been more efficient and optimized by deep learning based algorithms, where self-supervised feature extraction \cite{xue2022coarse} has been introduced. To extract global and local features from detected face, Weng \emph{et al.}\cite{weng2021attentive} implement a multi-branch network.  Xue \emph{et al.}\cite{xue2021transfer} build a relation-aware local-patch representations using retention module to explore extensive relations on the manifold local features based on multi-head self-attention and Transformer frameworks. Recently, both Li \emph{et al.} \cite{li2018occlusion} and Wang \emph{et al.}\cite{wang2020region} suggest attention networks based on regions to extract discriminative features which outperform for robust pose and occlusion-aware FEL.
\subsection{Transformers in FEL}
According to recent works \cite{naseer2021intriguing}, ViT \cite{dosovitskiy2020image} exhibits remarkable resilience against severe disruption and occlusion. To handle the shortcomings of FEL, such as poor quality samples, different backdrops and annotator’s various contexts, Li \emph{et al.} \cite{li2021mvt} introduce Mask Vision Transformer (MVT) to provide a mask that can remove complicated backdrops and facial occlusion, as well as an adaptive relabeling mechanism to fix inaccurate labels in real-world FEL archives. Addressing the inadequate performance of Transformers in recognizing the subtlety of expression in videos, Liu \emph{et al.} \cite{liu2023expression} develop a novel Expression Snippet Transformer (EST) which successfully models minuscule intra/inter snippet visual changes and effectively learns the long-range spatial-temporal relations. While Transformers have proven to perform well in FEL tasks, it still has vulnerabilities when dealing with multimodal (2D + 3D) datasets, as it needs more data. To combat this problem,  Li \emph{et al.} \cite{li2021mfevit} create a resilient lightweight multimodal facial expression vision Transformer for multimodal FEL data. Hwang \emph{et al.} \cite{hwang2022vision} propose Neural Resizer, a method that helps Transformers by balancing the noisiness and imbalance through data-driven information compensation and downscaling. Zhang \emph{et al.} \cite{zhang2022transformer} develop a Transformer-based multimodal information fusion architecture that utilizes dynamic multimodal features to fully leverage emotional knowledge from diverse viewpoints and static vision points.
\subsection{Uncertainty in FEL}
 Uncertainties refer to mainly cryptic expressions, imprecise and conflicting annotations in FEL tasks. To eliminate uncertainty annotations, Fan \emph{et al.} \cite{fan2020learning} introduce mid-level representation enhancement (MRE) and graph-embedded uncertainty suppressing (GUS). For performing both cleaning noisy annotations and classifying facial images, Viet \emph{et al.} \cite{lei2022mid} introduce a multitasking network architecture and Wen \emph{et al.} \cite{wen2016discriminative} implement center loss for applying intra-class compactness to extract discriminative features to reduce uncertain predictions. For determining facial expression with noisy annotations, Max-Feature-Map activation function is adopted by Wu \emph{et al.} \cite{wu2018light}. 

 \subsection{Attention Networks in FEL}
 Despite FEL gaining widespread recognition in the field of computer vision, two major problems- pose alterations and occlusion have not been properly addressed in terms of automatic expression detection. Wang et al. \cite{wang2020regiona} develop a system where Facial Expression Recognition (FER) datasets are annotated to include pose and real-world occlusion features. They also preset a Region Attention Network to encapsulate face areas and poses in an adaptive manner. Furthermore, a region-based loss function is introduced to induce larger attention weights. Another work that deals with such issues is by Zhao et al. \cite{zhao2021learning} where they propose a global multi-scale and local attention network (MA-Net) which consists of a multi-scale based local attention component with a feature pre-extractor that aids in focusing more on salient details. Inter-class similarity and models lacking high-order interactions among local aspects are some other problems present in current FER systems. Wen et al. present Distract your Attention Network (DAN) \cite{wen2021distract} with integral sections. They are Feature Clustering Network (FCN) which retrieves robust features by using a large-margin learning goal, Multi-head cross Attention Network (MAN) initializes a variety of attention heads to focus on multiple facial features at the same time and develop attention maps on these areas, and Attention Fusion Network (AFN) which integrates multiple attention maps from various regions into a single, unified map. Fernandez et al. \cite{marrero2019feratt} propose Facial Expression Recognition with Attention Net (FERAtt) which utilizes Gaussian space representation to recognize facial expressions. The architecture focuses on facial image correction, which employs a convolutional-based feature extractor in combination with an encoder-decoder structure, and facial expression categorization, which is in charge of getting an embedded representation and defining the facial expression.

 \section{Approach}
In our comprehensive approach, we propose a rigorous feature extraction strategy that is supported by ViT with reliability balancing mechanism  to tackle the difficulties of FEL. Its cutting-edge framework is composed of a variety of components that function together to provide solutions that are accurate and reliable.
We start by scaling the input photos before initiating the augmentation procedure in order to achieve better augmentation. Following image scaling, there is rotation, color enhancement, and noise balancing. The images are then randomly cropped for the optimal outcomes after extensive augmentation. Our pipeline meticulously addresses different biases and overfitting possibilities that may exist in the training data by randomly selecting a few images from each video and assembling them. Furthermore, we randomly select a set of images representing each expression for every epoch. The overall selection process and the parameters varies on different datasets based on their distribution of labels, number of classes and images.

In our approach, the cross-attention ViT is used in the feature extraction process. Which is aimed at tackling common FEL issues, including scale sensitivity, intra-class discrepancy, as well as inter-class similarity. We employ a pre-trained landmark extractor to locate different facial landmarks on a particular face. Afterward, we use a pre-trained image backbone model to extract features from the image accordingly. We utilize multiple feature extractors to detect low-level to high-level features in the image using different facial landmarks. After feature extraction, collected multi-level feature information is integrated. We use a cross-attention mechanism for linear computation and it integrates multi-level features and provides feature vector embedding using a optimised version of POSTER++ \cite{mao2023poster}. This comprehensive feature extraction framework provides correctness and dependability in the final output vector of length $768$ by combining a cross-attention mechanism with substantial feature extraction. We also use an additional linear reduction layer to decrease the feature vector size to $128$. Primary label distributions are generated using logits resulted by Multi-Layer Perceptrons. Multi-Layer Perceptrons (MLP) include variety of hidden layers that enable them to process information with great precision and accuracy.  We calculated confidence value based on the primary label distribution using Normalized Entropy to evaluate the reliability\cite{le2023uncertainty} of these models. 

We introduce a novel reliability balancing method to solve the limitations of modern FEL models. Modern FEL models still have several limitations, especially when it comes to making precise predictions for classes because their images are quite similar. This issue leads to a biased and unreliable model. Our reliability balancing method can increase the prediction capability of unbalanced and  erroneous predictions, thereby improving the performance of the model. We achieve enhanced reliability and confidence by placing multiple learnable anchors in the embedding space and using multi-head self- attention mechanism, which help identify the closest neighbors of erroneous predictions to improve the prediction ability. The use of anchor spaces and attention values have proven to be highly effective in stabilizing label distribution, thereby resulting in better performance overall. We obtain additional regularization whenever possible by implementing dropout layers for more robustness. Our approach ensures that the model is resilient even in the circumstance of noisy or inadequate data; minimizing possible bias and overfitting. The resulting model integrating extensive feature extraction with reliability balancing, is remarkably precise and able to make credible predictions even in the context of ambiguity. The overall pipeline is illustrated in Fig. \ref{fig:2}.

\subsection{Problem Formulation}
Let ${x}^i$ be the $i$-th instance variable in the input space $\mathcal{X}$ and $y^i \in \mathcal{Y}$ be the label of the $i$-th instance with $\mathcal{Y} = \{y_1, y_2 \dots y_N\}$ being the label set.
Let $\mathcal{P}^n$ be the set of all probability vectors of size $n$.
Furthermore, let ${l}^i \in \mathcal{P}^N$ be the discrete label distribution of $i$-th instance.
Additionally, let ${e} = p({x; \theta_p})$ be the embedding output of the Window-Based Cross-Attention ViT (explained in \ref{Window-Based Cross-Attention ViT}) network $p$ with parameters ${\theta_p}$
and let $f({e}; {\theta_f})$ be the logit output of the MLP classification head network $f$ with parameters ${\theta_f}$.

\subsection{Window-Based Cross-Attention ViT} \label{Window-Based Cross-Attention ViT}
We use a complex image encoder to capture distinctive patterns from the input images. We obtain feature embedding vectors in our proposed pipeline with a refined and optimised version of POSTER++\cite{mao2023poster}, a window-based cross-attention ViT network.

We employ a window-based cross-attention mechanism to achieve linear computation. We extract features by the image backbone and facial landmark detectors. We used IR50 \cite{wang2021face} as image backbone and MobileFaceNet \cite{chen2018mobilefacenets} as facial landmark detector. For each level, firstly, division of image features $X_{img} \in \mathcal{R}^{N \times D}$ is performed, where $N$ represents the number of classes and $D$ denotes the feature dimensions. These divided image features are transformed into many non-overlapping windows,  $z_{img}\in \mathcal{R}^{M \times D}$ where $z_{img}$ contains \emph{M} tokens. Then, down-sampling of the landmark feature $X_{lm} \in \mathcal{R}^{C \times H \times W}$ takes place, where $C$ is the number of channels in the attention network, $H$ and $W$ are the height and width of the image. The down-sampled features are converted into the window size, where the smaller representation of the image is taken and it is represented by $z_{lm}\in \mathcal{R}^{c \times h \times w}$ where $c = D, h \times w$ = M. The features are reshaped in accordance with $z_{img}$’s shape.
The cross-attention with \emph{I} heads in a local window can be formulated as follows at this point:

\begin{equation}
    q = z_{lm}w_{q}, k = z_{img}w_{k}, v = z_{img}w_{v}
\end{equation} 
\begin{equation}
    o^{(i)} = softmax(q^{(i)}k^{(i)T}/\sqrt{d}+b)v^{(i)}, i = 1,\ldots,I
\end{equation}
\begin{equation}
      o = [o^{(1)},\ldots,o^{(I)}]w_o
\end{equation}

where $w_{q}$, $w_{k}$, $w_{v}$ and $w_{o}$ are the matrices used for mapping the landmark-to-image features, and $q, k, v$ denote the query matrix for landmark stream, and key, and value matrices for the image stream, respectively from different windows used in the window-based attention mechanism. [·] represents the merge operation where the images patches are combined to identify the correlations between them and lastly, the relative position bias is expressed as $b \in \mathcal{R}^{I \times I}$ which aids in predicting the placement between landmarks and image sectors.

We use the equations above for calculating the cross-attention for all the windows. This method is denoted as \textbf{W}indow-based \textbf{M}ulti-head \textbf{C}ros\textbf{S}-\textbf{A}ttention (W-MCSA). Using the equations below, the Transformer encoder for the cross-fusion can be calculated as:

\begin{equation}
    X'_{img} = {W\textrm{-}MCSA}_{(img)}+X_{img}
\end{equation}
\begin{equation}
    X_{img\_O} = MLP(Norm(X'_{img}))+X'_{img}
\end{equation}

where $X'_{img}$ is the combined image feature using W-MCSA, $X_{img\_O}$ the output of the Transformer encoder, and $Norm$(·) represents a normalization operation for the full image of all windows combined. Using window information and dimensions ($z_{img}, M, D, C, H, W, etc.$), we extract and combine window based feature information to $Xo_i$ ($i$-th level window-based combined features of each image) from $X_{img\_O}$ (extracted features of all windows of each image together).

We introduce a Vision Transformer to integrate the obtained features at multiple scales $Xo_1,...,Xo_i$. Our attention mechanism is able to capture long-range dependencies as it combines information tokens of all scale feature maps like POSTER++\cite{mao2023poster}. The method is described as:

\begin{equation}
    Xo = [Xo_1,...,Xo_i]
\end{equation}
\begin{equation}
    Xo' = MHSA(Xo) + Xo
\end{equation}
\begin{equation}
    Xo_{out} =MLP(Norm(Xo)) + Xo'
\end{equation}
where, [·] denotes concatenation, $MHSA$(·) denotes multi-head self-attention mechanism, $MLP$(·) is the multi-layer perceptron.

\begin{figure}[pht] 
\centering {\includegraphics[scale=0.4]{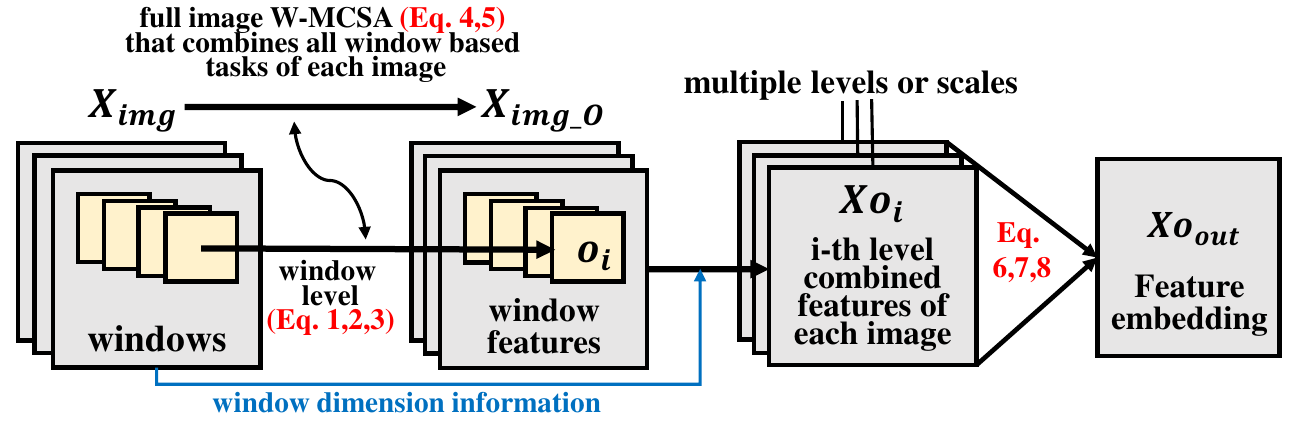}}
\caption{Data flow in the Window-Based Cross-Attention ViT network}\label{fig:WC_ViT}
\end{figure}
Output of the multi-scale feature combination module $Xo_{out}$, which is equal to feature embedding $e$, is the final output of the encoder network denoted by $p({x; \theta_p})$.

\subsection{Reliability Balancing} \label{Reliability Balancing with Anchors}
Majority of Facial Expression Learning datasets are labeled using only one label for each sample. Inspired by \cite{le2023uncertainty,deng2019arcface}, we provide an alternative approach, in which, we learn and improve label distributions utilizing a label correction approach. We calculate a label distribution primarily that uses the embedding ${e}$ directly into the MLP network. Subsequently, the reliability balancing section employs label correction techniques to stabilize the primary distribution. This results in improved predictive performance through more accurate and reliable labeling. \\

\textbf{Primary Label Distribution. }
From sample ${x}$, using the $p$ network we can generate
the corresponding embedding ${e} = p({x; \theta_p})$ and using the
$f$-network consisting MLP, we can generate the corresponding discrete primary label distribution:
\begin{equation}
{l} = softmax(f({e; \theta_f}))
\end{equation}
We use the information contained in the label distribution with label corrections during training to improve the model performance. 

\textbf{Confidence Function. }
To evaluate the credibility of predicted probabilities, a confidence function is designed. Let $C: \mathcal{P}^N \to [0, 1]$, be the confidence function. 
$C$ measures the certainty of a prediction made by the classifier using normalized entropy function H({l}). The functions are defined as:

\begin{equation}
    C({l}) = 1 - H({l})
\end{equation}
\begin{equation}
    H({l}) = -\frac{\sum_i l^i \log(l^i)}{N}
\end{equation}

For a distribution where all probabilities are equal; normalized entropy is 1, and confidence value is 0. Where one value is equal to 1, and others are equal to 0; normalized entropy is 0, and confidence value is 1.

\subsection{Label Correction}
The conundrum of label accuracy, distribution stability, and reliability has been a mainstream problem in FEL. The novel approach we propose to resolve this is a combination of two distinct measures of label correction: Anchor Label Correction and Attentive Correction. By leveraging geometric similarities and state-of-the-art multi-head attention mechanism, we are designing predicted labels are not only accurate but also stable and reliable.

\subsubsection{Anchor Label Correction}

\textbf{Anchor Notations. } 
We define anchor ${a}^{i,j}$ $(i \in \{1, 2 \dots, N\}, j \in \{1, 2 \dots K\}$)
to be a point in the embedding space.
Let $\mathcal{A}$ be a set of all anchors.
During training we use $K$ trainable anchors for each label, with $K$ being a hyperparameter.
We assign another label distribution ${m}^{i,j} \in \mathcal{P}^N$ to anchor ${a}^{i,j}$,
where ${m}^{i,j}$ is defined as:
$$
m^{i,j}_k =
\begin{cases}
    1, \text{ if } k = i \\
    0, \text{ otherwise}\\
\end{cases}
$$
Intuitively, here it means anchors ${a}^{1, 1}, {a}^{1, 2} \dots {a}^{1, K}$ are labeled as
belonging to class 1, anchors ${a}^{2, 1}, {a}^{2, 2} \dots {a}^{2, K}$ are labeled as
belonging to class 2 and so on.\\

\textbf{Geometric Distances and Similarities. } 
To correct the final label and stabilize the distribution we use the geometric information about similarity between the embeddings and a fixed number of learnable points in the embedding space called anchors.

The similarity score $s^{ij}({e})$
is a normalized measure of similarity between an embedding ${x}$ and
an anchor ${a}^{ij} \in \mathcal{A}$.

The distance between embedding ${e}$ and anchor ${a}$ for each batch and class is defined as:

\begin{equation}
    d({e}, {a}) = \sqrt{\sum_{dim_{e}} {|{a} - {e}|^2}}
\end{equation}

Here, $dim_{e}$ is the dimension of embedding ${e}$. Distances $|{a} - {e}|^2$ are reduced over the last dimension $dim_{e}$ and element--wise square root is taken for stabilizing values.

The similarity score $s^{ij}$ is then obtained by normalizing distances as softmax:
\begin{equation}
    s^{ij}({e}) = \frac{\exp(-\frac{d({e}, {a}^{ij})}{\delta})}{\sum_i^N \sum_j^K \exp(-\frac{d({e}, {a}^{ij})}{\delta})}
\end{equation}
where, $\delta$ is a hyperparameter, which is used in the computation of softmax to control the steepness of the function. The default value used for $\delta$ is 1.0.\\

\textbf{Correction. } 
From similarity scores we can calculate the anchor label correction term as follows:
\begin{equation}
    {t_{g}}({e}) = \sum_i^N \sum_j^K s^{ij}({e}) {m}^{ij}
\end{equation}

\subsubsection{Attentive Correction}
\textbf{Multi-head Self-Attention.} For multi-head attention \cite{vaswani2017attention}, Let a query with query embeddings $q \in \mathcal{R}^{d_{Q}}$, key embeddings $k \in \mathcal{R}^{d_{K}}$, and value embeddings $v \in \mathcal{R}^{d_{V}}$ is given. With the aid of independently learned projections, they can be modified with $h$, which is the attention head. These parameters are then supplied to attention pooling. Finally, these outputs are altered and integrated using another linear projection. The process is described as follows:

\begin{equation}
h_i = f(W_i^{(q)}q, W_i^{(k)}k, W_i^{(v)}v)\in \mathcal{R}^{p_{V}},
\end{equation}
where $W_i^{(Q)} \in \mathcal{R}^{d_Q \times p_Q}, W_i^{(K)} \in \mathcal{R}^{d_K \times p_K}, W_i^{(V)} \in \mathcal{R}^{d_V \times p_V}$ are trainable parameters, and $f$ is the attentive pooling and $h_i(i = 1,2,...,n_{heads})$ is the attention head. The output obtained through learnable features, $W_{out} \in \mathcal{R}^{p_{out} \times hp_{out}}$ can be categorized as:

\begin{equation}
W_{out}\begin{bmatrix}h_1 \\ \vdots \\ h_{n_{heads}} \end{bmatrix}\in\mathcal{R}^{p_{out}}
\end{equation}
As we are using self-attention, all inputs ($q, k, v$ denoting query, key and value parameters respectively) are equal to the embedding ${e}$.

\textbf{Correction.} To additionally correct and stabilize the label distributions, we use attention-based similarity function.
The embedding ${x}$ is passed through the multi-head self-attention layer to obtain attentive correction term ${t_{a}}$. General formula:
\begin{equation}
t_{a} = softmax(W_{out})
\end{equation}
${t_{a}}$ is reshaped as required.

\subsubsection{Final Label correction}
To combine the correction terms, we use weighted sum, with weighting being
controlled by the confidence of label corrections.

\begin{equation}
{t} = \frac{c_g}{c_g + c_a} {t_g} + \frac{c_a}{c_g + c_a} {t_a}
\end{equation}
where $c_g = C({t_g})$ and $c_a = C({t_a})$.

Finally, to obtain the final label distribution $L_{final}$, we use a weighted sum
of label distribution ${l}$ and label correction ${t}$.
\begin{equation}
    {L_{final}} = \frac{c_l}{c_l + c_t} {l} + \frac{c_t}{c_l + c_t} {t}
\end{equation}
where $c_l = C({l})$ and $c_t = C({t})$.\\

The label with maximum value in final corrected label distribution $L_{final}$ is provided as corrected label or final predicted label.

\subsection{Loss Function}
Loss function used to train the model consists of three terms such as class distribution loss, anchor loss, and center loss. 

\textbf{Class Distribution Loss ($\mathcal{L}_{cls}$):}
To make sure each example is classified correctly, we use the negative log-likelihood loss between the corrected label distribution ${L^i}$ and label ${y^i}$:
\begin{equation}
\mathcal{L}_{cls} = - \sum_i^m \sum_j^N y^i_j \log L^i_j
\end{equation}

\textbf{Anchor Loss ($\mathcal{L}_{a}$):}
In order to amplify the discriminatory capacity of the model, we want to make margins between anchors large so that we add an additional loss term:
\begin{equation}
\mathcal{L}_{a} = - \sum_i \sum_j \sum_k \sum_l |{a}^{ij} - {a}^{kl}|^2_2
\end{equation}
We include the negative term in front because we want to maximize this loss. The loss is also normalized for standard uses.

\textbf{Center Loss ($\mathcal{L}_{c}$):}
To make anchors good representation of their class, we want to make sure anchors and embeddings of the same class stay close in the embedding space.
To ensure that, we add an additional error term:
\begin{equation}
\mathcal{L}_c = \min_{k} |{x}^i - {a}^{y^ik}|^2_2
\end{equation}

\textbf{Total Loss ($\mathcal{L}_{total}$):}
Our final loss function can be defined as:
\begin{equation}
    \mathcal{L}_{total} = \lambda_{cls} \mathcal{L}_{cls}  + \lambda_{a} \mathcal{L}_a + \lambda_{c} \mathcal{L}_c
\end{equation}
with $\lambda_{cls}, \lambda_{a}, \lambda_{c}$ being hyperparameters, used to keep the loss functions in same scale.

\section{Experiments}
\subsection{Datasets} \label{Datasets}
 
\textbf{Aff-Wild2} \cite{kollias2023abaw1} has 600 annotated audiovisual videos with 3 million frames for categorical and dimensional affect and action units models. It has 546 videos for expression recognition, with 2.6 million frames and 437 subjects. Experts annotated it frame-by-frame for six distinct expressions - surprise, happiness, anger, disgust, fear, neutral expression, and 'other' emotional states.

\textbf{RAF-DB} \cite{li2017reliable, li2019reliable} contains a whopping 30,000 distinct images that encompass 7 unique labels for expressions. The images have been meticulously labeled by 40 individual annotators and boast a diverse range of attributes, including subsets of emotions, landmark locations, bounding boxes, race, age range, and gender attributions.

\textbf{JAFFE} \cite{lyons2020coding, lyons2021excavating} dataset comprises facial expressions from ten Japanese women, featuring seven prearranged expressions and multiple images for each individual depicting the respective expressions. The dataset encompasses a total of 213 images, with each image rated on six facial expressions by 60 Japanese viewers. The images are stored in an uncompressed 8-bit grayscale TIFF format, exhibit a resolution of 256x256 pixels, and contain no compression.

\textbf{FERG-DB} \cite{aneja2016modeling} database is yet another benchmark repository that features 2D images of six distinct, stylized personages that were produced and rendered using the sophisticated MAYA software. Comprising an impressive quota of 55,767 meticulously annotated facial expression images, this mammoth database is categorized into seven distinct emotion classes: disgust, surprise, sadness, fear, joy, neutral, and anger.

\textbf{FER+} \cite{BarsoumICMI2016} annotations have bestowed a novel set of labels upon the conventional Emotion FER dataset. Each image in FER+ has undergone rigorous scrutiny from a pool of 10 taggers, thereby generating high-quality ground truths for still-image emotions that surpass the original FER labels. It contains seven main labels and one for exclusions.

\begin{figure}[pht] 
\centering {\includegraphics[scale=0.39]{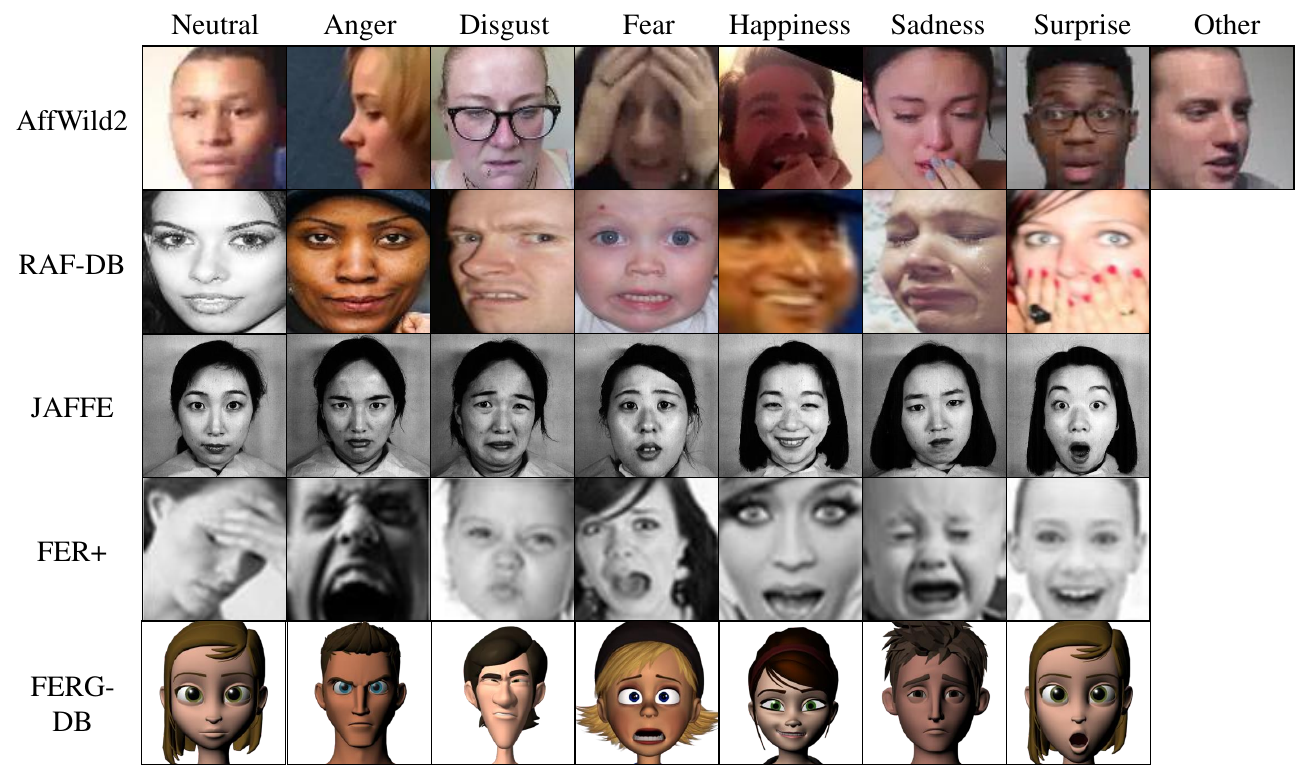}}
\caption{Examples of training samples in different datasets}\label{fig:Datasets}
\end{figure}

\subsection{Data Distribution Adjustments}
\subsubsection{Augmentation}
Sample augmentation entails artificially amplifying the training set by crafting altered replicas of a dataset employing extant data, expediting the discernment of significant attributes from the data. This technique is also particularly useful for achieving robust training data. In a typical FEL problem, the usual data preprocessing and augmentation steps include image resizing, scaling, rotating, padding, flipping, cropping, color augmentation and image normalization. 

In this study, we are utilizing multiple image processing techniques, including Image Resizing and Scaling, Random Horizontal Flip, Random Crop, etc. 

\textbf{Image Resizing and Scaling.}
Bi--linear interpolation is used as image resizing method. It uses linear interpolation in both directions to resize an image. This process is repeated until the final result is achieved \cite{press2007numerical}.\\
Suppose we seek to evaluate the function $f_{s}$ at the coordinates $(x, y)$. We possess the function value of $f_{s}$ at the quadrilateral vertices $Q_{11}=\left(x_1, y_1\right), Q_{12}=\left(x_1, y_2\right), Q_{21}=\left(x_2, y_1\right)$, and $Q_{22}=\left(x_2, y_2\right)$. 
Here is the equation for interpolation,
\begin{equation}
\begin{aligned}
& f_{s}(x, y)= \frac{1}{\left(x_2-x_1\right)\left(y_2-y_1\right)}
\left[\begin{array}{ll}x_2-x & x-x_1 \end{array}\right]\\ 
& \qquad \left[\begin{array}{ll}
f_{s}\left(Q_{11}\right) & f_{s}\left(Q_{12}\right) \\
f_{s}\left(Q_{21}\right) & f_{s}\left(Q_{22}\right)
\end{array}\right]\left[\begin{array}{l}
y_2-y \\
y-y_1 \end{array}\right]
\end{aligned}
\end{equation}
The operation is executed on all necessary pixels until the complete image is suitably rescaled.\\

\textbf{Random Horizontal Flip.}
The random flip operation encompasses an arbitrary flipping of the input image with a designated probability. Consider $A \in R^{m \times n}$ to be the provided torch tensor. The matrix $A=A_{i j}$, where $i \in\{1, \ldots, m\}$ signifies the row and $j \in\{1, \ldots, n\}$ corresponds to the column of the image. Horizontal Flip equation $\Longrightarrow A_{i(n+1-j)}$. This operation exchanges the columns of $A$ such that the initial column of $A$ corresponds to the final column of $A_{i(n+1-j)}$, while the ultimate column of $A$ is equivalent to the first column of $A_{i(n+1-j)}$.\\

\textbf{Random Crop.}
Random cropping entails excising a section of the input image at a serendipitous location. A torch tensor image is envisaged to possess a shape of […, H, W], where …… denotes any number of antecedent dimensions. In cases where non-static padding is implemented, the input should comprise at most two introductory dimensions.

\subsubsection{Data Refinement}
Inconsistent class distribution can cause bias and overfitting. Some datasets may have excess data on certain faces, introducing bias towards specific facial data. Providing the model with equally distributed information from all classes and faces can help it accurately distinguish between them and avoid overfitting on common data. Additionally, refining the training datasets can ensure the effective distribution of classes, preventing biases and enhancing the model's performance. 

This refinement process is initiated during every epoch, each of which entails a unique set of data for training purposes. Our methodology for refining data comprises two sections. During the training stage, a total of \(N\) preprocessed, cropped, and aligned images are selected in a stochastic manner from each video or group of faces, based on the dataset. These images are then aggregated into a pool, from which \(M\) images per expression are randomly selected for training purposes. Thus, a batch of ($M \times$ number of classes) images is assembled randomly during every epoch, ensuring balanced training information and counteracting potential biases and overfitting.

\begin{table*}[pht]
\centering
\caption{Number of Anchors \emph{K} vs. Accuracy (\%) ($\uparrow$) means increase in accuracy. [KEY: \textcolor{red}{\textbf{Best}}
*= Our evaluation of the proposed model]} 
\begin{tabular}{c c c c c c c c c c c}
\hline
\textbf{K}             & 0     & 1                  & 2                  & 3                  & 4  & 5             & 6                 & 8                 & 10    & 20    \\ \hline
\textbf{Accuracy (\%)} & 68.92 & +1.82 ($\uparrow$) & +2.03 ($\uparrow$) & +2.11 ($\uparrow$) & +2.19 ($\uparrow$) & +2.24 ($\uparrow$)& +2.26 ($\uparrow$)&\textcolor{red} {\textbf{+2.29}} ($\uparrow$) & \textcolor{red}{\textbf{+2.29}} ($\uparrow$)& +0.51 ($\uparrow$)\\ \hline
\label{table:Number of Anchors K vs. Accuracy}
\end{tabular}
\end{table*}

\begin{table*}[pht]
\centering
\caption{\emph{K} for Different Noise vs. Accuracy (\%) ($\uparrow$).  [KEY: \textcolor{red}{\textbf{Best}}
*= Our evaluation of the proposed model]} 
\begin{tabular}{c c c c c c c c c c c}
\hline
\emph{K} &  \multicolumn{10}{c}{Noise} \\ \hline
         & 0             & 5             & 10             &  15             & 20             & 25             & 30             & 35             & 40             & 50             \\ \hline
0 & 68.92 & 68.41 & 67.7 & 63.69 & 54.99 & 50.36 & 41.18 & 36.94 & 34.12 & 30.92 \\ \hline
1 & 70.74 & 70.52 & 70.02 & 66.51 & 59.81 & 51.18 & 44.00 & 37.76 & 35.94 & 33.79 \\ \hline
2 & 70.95 & 70.61 & 70.23 & 66.72 & 60.02 & 51.39 & 44.21 & 37.97 & 36.15 & 34.00 \\ \hline
3 & 71.03 & 70.64 & 70.31 & 66.80 & 60.10 & 51.47 & 44.29 & 38.05 & 36.23 & 34.08 \\ \hline
4 & 71.11 & 70.65 & 70.39 & 66.88 & 60.18 & 51.55 & 44.37 & 38.13 & 36.31 & 34.16 \\ \hline
5 & 71.16 & 70.70 & 70.44 & 66.93 & 60.23 & 51.60 & 44.42 & 38.18 & 36.36 & 34.21 \\ \hline
6 & 71.18 & 70.71 & 70.46 & 66.95 & 60.25 & 51.62 & 44.44 & 38.20 & 36.38 & 34.23 \\ \hline
7 & 71.21 & 70.71 & 70.49 & 66.98 & 60.28 & 51.65 & 44.47 & 38.23 & 36.41 & 34.26 \\ \hline
8 & 71.24 & 70.72 & 70.52 & 67.01 & 60.31 & 51.66 & 44.49 & 38.26 & 36.43 & 34.29 \\ \hline
9 & 71.24 & 70.73 & 70.52 & 67.01 & 60.32 & 51.68 & 44.50 & 38.26 & 36.44 & 34.29 \\ \hline
10 & \textcolor{red}{\textbf{71.25}} & 70.73 & 70.53 & 67.02 & 60.33 & 51.69 & 44.51 & 38.27 & 36.45 & 34.3 \\
\hline
\label{table:Noise vs. Accuracy}
\end{tabular}
\end{table*}

\subsection{Implementation Details}
For each dataset, we take the cropped and aligned images, exclusively. We resize them to 256$\times$256 and took a random crop of 224$\times$224. To deal with overfitting and imbalance of data in particular categories of expression, we pre-process the data using heavy augmentation methods. For the data refinement, we consider 512 images for each video/face and then, the images are combined from all to create an unbiased set. For training, 500 images are taken for each class category from the set.

The images' embeddings are collected from the Cross Attention ViT network from images. Three loss functions are combined to train our model. Anchor loss is aimed to keep the anchors far apart from each other and center loss is aimed at the embeddings to be close to the anchors. Class distribution loss is used to classify the classes correctly. The number of epochs used is 1000 for training. In order to optimize our model, we utilize the ADAM optimizer algorithm with an inaugural learning rate of 0.0003, thereby ensuring the convergence of the gradient descent process towards a global minimum with methodical and efficacious precision. But, the learning rate is scheduled using exponential decay with $\gamma$ of 0.995. MLP consisting of 2 hidden layers of size 64 is used for primary prediction. Each layer is followed by a ReLU activation, a dropout layer, and a batch normalization layer except the last one.  The dropout layers have a drop probability of 0.5 for regularization.

\subsection{Evaluation Metric}
Throughout the experiments, accuracy is utilized as the primary evaluation metric, which is a fundamental concept that measures the correctness of predictions made by a model.

In the case of multi-class classification, accuracy measures the proportion of correct classifications ($n_{correct}$) and the total number of classified terms ( $n_{total}$). The equation is:
\begin{equation}
\text { Accuracy }=\frac{n_{correct}}{n_{total}}
\end{equation}

\subsection{Ablation Studies}
In order to demonstrate the efficacy of our approach, we undertake a series of ablation studies aimed at assessing the impact of critical parameters and components on the ultimate performance outcomes. The Aff-Wild2 dataset is utilized as the primary dataset throughout experimental procedures to enable a comprehensive evaluation of the effectiveness of our proposed method. It includes emotion information of 8 different classes.

\textbf{Number of Anchors \emph{K} vs. Accuracy.}
The presented findings in Table \ref{table:Number of Anchors K vs. Accuracy} reveal that the proposed approach attains optimal recognition accuracy when the number of anchors is set to a range of 8-10. The data shows a gradual rise in accuracy until reaching the certain range of \emph{K}, beyond which it experiences a sharp decline. A small number of anchors fail to effectively model expression similarities while an excessive number of anchors introduces redundancy and noise in the embedding space, resulting in a decline in performance. Subsequently, for our subsequent experiments, we have opted to set the number of anchors at 8-10.

\begin{table*}[pht]
\centering
\caption{\emph{K} for Different Label Smoothing Terms vs. Accuracy (\%) ($\uparrow$).  [KEY: \textcolor{red}{\textbf{Best}}
*= Our evaluation of the proposed model]} 
\begin{tabular}{c c c c c c c c c c c c c}
\hline
\emph{K} &  \multicolumn{12}{c}{Label smoothing Terms} \\ \hline
  & 0 & 5 & 10 & 11 & 15 & 18 & 20 & 25 & 30 & 35 & 40 & 50 \\ \hline
0 & 68.92 & 69.16 & 69.18 & 69.18 & 69.03 & 68.64 & 67.50 & 64.27 & 61.75 & 59.34 & 55.83 & 50.88 \\ \hline
1 & 70.74 & 71.38 & 71.94 & 71.97 & 71.46 & 70.68 & 70.62 & 67.59 & 63.07 & 60.66 & 56.15 & 51.20 \\ \hline
2 & 70.95 & 71.59 & 72.15 & 72.18 & 71.67 & 70.89 & 70.83 & 67.80 & 63.28 & 60.87 & 56.36 & 51.41 \\ \hline
3 & 71.03 & 71.67 & 72.23 & 72.26 & 71.75 & 70.97 & 70.91 & 67.88 & 63.36 & 60.95 & 56.44 & 51.49 \\ \hline
4 & 71.11 & 71.75 & 72.31 & 72.34 & 71.83 & 71.05 & 70.99 & 67.96 & 63.44 & 61.03 & 56.52 & 51.57 \\ \hline
5 & 71.16 & 71.80 & 72.36 & 72.39 & 71.88 & 71.10 & 71.04 & 67.01 & 63.49 & 61.08 & 56.57 & 51.62 \\ \hline
6 & 71.18 & 71.82 & 72.38 & 72.41 & 71.90 & 71.12 & 71.06 & 67.03 & 63.51 & 61.10 & 56.59 & 51.64 \\ \hline
7 & 71.21 & 71.85 & 72.41 & 72.44 & 71.93 & 71.15 & 71.09 & 67.06 & 63.54 & 61.13 & 56.62 & 51.67 \\ \hline
8 & 71.24 & 71.86 & 72.44 & 72.47 & 71.96 & 71.18 & 71.12 & 67.09 & 63.57 & 61.16 & 56.65 & 51.70 \\ \hline
9 & 71.24 & 71.88 & 72.44 & 72.47 & 71.96 & 71.18 & 71.12 & 67.09 & 63.57 & 61.16 & 56.65 & 51.70 \\ \hline
10 & 71.25 & 71.89 & 72.45 & \textcolor{red}{\textbf{72.48}} & 71.97 & 71.19 & 71.13 & 67.1 & 63.58 & 61.17 & 56.66 & 51.71 \\ \hline
\label{table:Different Label Smoothing Terms vs. Accuracy}
\end{tabular}
\end{table*}
\textbf{\emph{K} for Different Noise vs. Accuracy.}

The presented findings in Table \ref{table:Noise vs. Accuracy} demonstrate that as we increase the level of noise, our model's accuracy decreases. This reduction in accuracy is due to the impact of noise on the clarity and complexity of the data, which makes it harder for our model to make accurate predictions. However, we can improve our model's performance by increasing the value of K, which enables our model to take into account more neighboring points when making decision boundaries, thus reducing the influence of noisy or outlier data points.

In particular, we observe that as we increase K, there is a modest but consistent improvement in accuracy for each level of noise. For example, at a noise level of 20, our model's accuracy improves from 54.99\% with K=0 to 60.33\% with K=10, representing a significant improvement of 5.34\% points. However, we should bear in mind that increasing K also increases computational complexity and memory usage, so we need to find a balanced tradeoff between model accuracy and practical considerations. We should also avoid excessively high values of K, which may lead to oversmoothing and loss of detail in the classification decision boundaries.

\textbf{\emph{K} for Different Label Smoothing Terms vs. Accuracy.}
Table \ref{table:Different Label Smoothing Terms vs. Accuracy} demonstrates the influence of label smoothing on model accuracy at various K settings. When smoothing terms from 0 to 50 are examined, accuracy measurements are provided as percentages. Regardless of the smoothing term used, the findings show an evident pattern where an increase in K values typically results in an improvement in model accuracy. Nonetheless, depending on the smoothing term used, this improvement varies in scope. \\
When examining the data, we found that the baseline model's accuracy is 68.92\% without smoothing. As K increases, accuracy increases in steps, peaking at 71.25\% when K=10. Applying smoothing methods also follows a similar pattern, with maximum accuracy of 71.80\% at smoothing term = 5. More smoothing leads to decreased accuracy. With K=10, max accuracy of 71.89\% is seen at smoothing term = 5, declining to a minimum of 51.20\% at smoothing term = 40. Smoothing terms from 5 to 20 have quite similar accuracy values, making 10 and 11 reasonable options to reduce overconfidence while discovering significant patterns. We conclude that a smoothing term of 11 is the best option for our model considering all relevant aspects.

\textbf{Different Anchor Loss Setups vs. Performance.}
From table \ref{tab:Different Anchor Loss Setups}, we observe how the anchor loss adjustment hyperparameter ($\lambda_{a}$) affects the model's performance.
\begin{table}[h]
\centering
\caption{Different Anchor Loss Setups ($\lambda_{a}$) vs. Performance (\%) ($\uparrow$). [KEY: \textcolor{red}{\textbf{Best}}
*= Our evaluation of the proposed model]}
\label{tab:Different Anchor Loss Setups}
    \begin{tabular}{l c c c c }
        \hline
        $\lambda_{a}$ & Precision & Accuracy & F1 Score  & Recall\\
        \hline
        Ideal (x1) & \textcolor{red}{\textbf{90.7}} & \textcolor{red}{\textbf{89.6}} & \textcolor{red}{\textbf{89.4}} & 89.1 \\ \hline
        Higher (x100) & 41.3 & 47.9 & 38.7 & 47.0 \\ \hline
        Lower (x0.01) & 80.5 & 82.3 & 80.5 & 81.1 \\ \hline
        No Effect (x0) & 87.3 & 88.5 & 88.4 & \textcolor{red}{\textbf{90.3}} \\ \hline
    \end{tabular}
\end{table}

 The ideal configuration has the highest precision (90.7\%), accuracy (89.6\%) and F1 Score (89.4\%) but a lower recall (89.1\%) than the maximum (90.3\%). Higher emphasis on $\lambda_{a}$ decreases model performance, but lower emphasis keeps it quite stable. When the anchor loss is not applied ($\lambda_{a} = 0$) in label correction, it results in higher recall because it focuses on cross-entropy more, but all the other metrics fall due to imbalance in distribution.

\begin{figure*}[p] 
\centering {\includegraphics[scale=0.2]{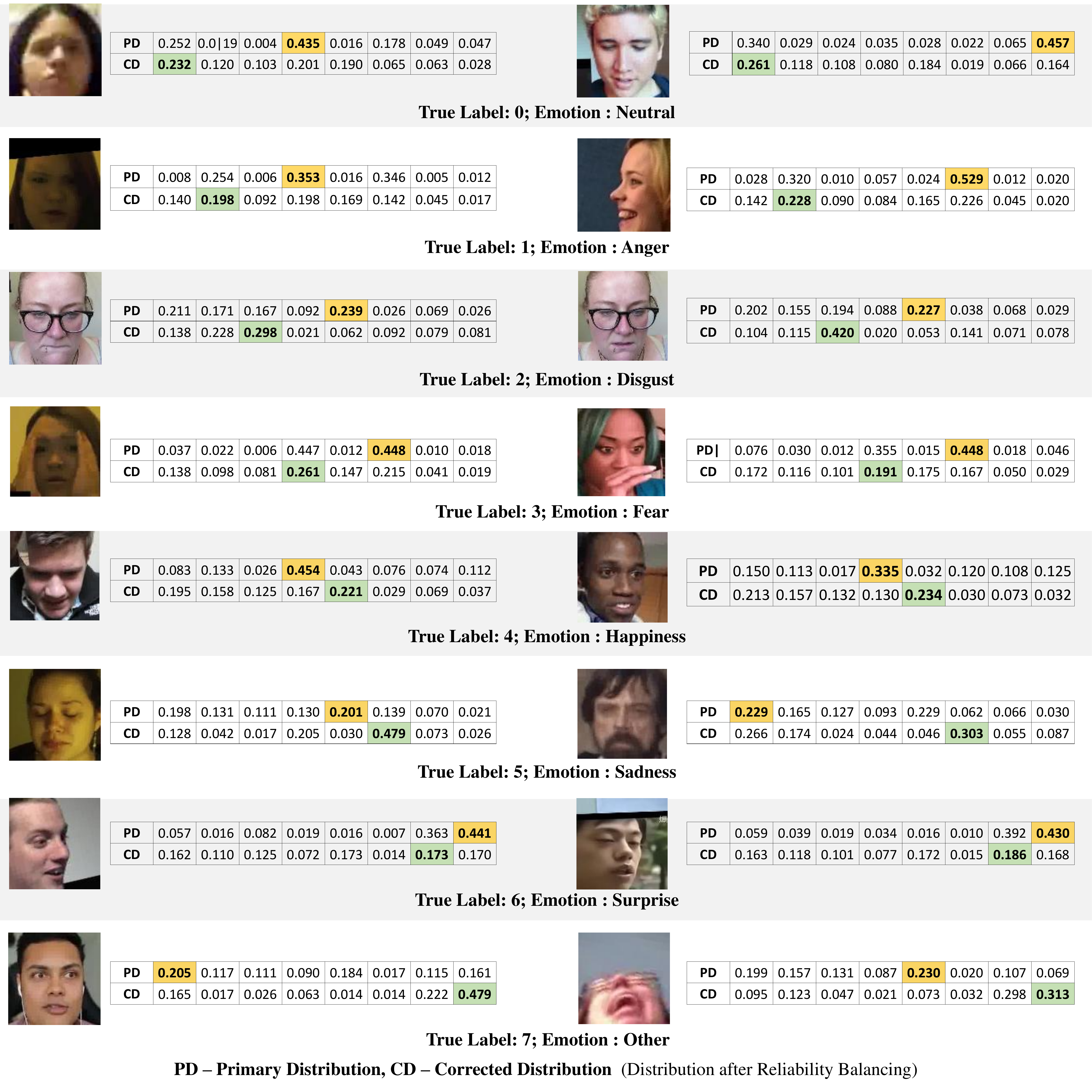}}
\caption{Observation of \textbf{confidence probability distributions in ARBEx} using \textit{Aff-Wild2} dataset. Eight different emotions—\textbf{Neutral, Anger, Fear, Disgust, Happiness, Sadness, Surprise,} and \textbf{Other}—are represented by columns under each image sequentially. \textbf{Primary Distribution (PD)} is the initial prediction while \textbf{Corrected Distribution (CD)} is the accurate prediction after \textbf{Reliability Balancing}. The correct label after reliability balancing is marked as green, and the inaccurate primary prediction label is marked as yellow.
}\label{fig:3}
\end{figure*}

\begin{figure*}[pht] 
\centering {\includegraphics[scale=0.55]{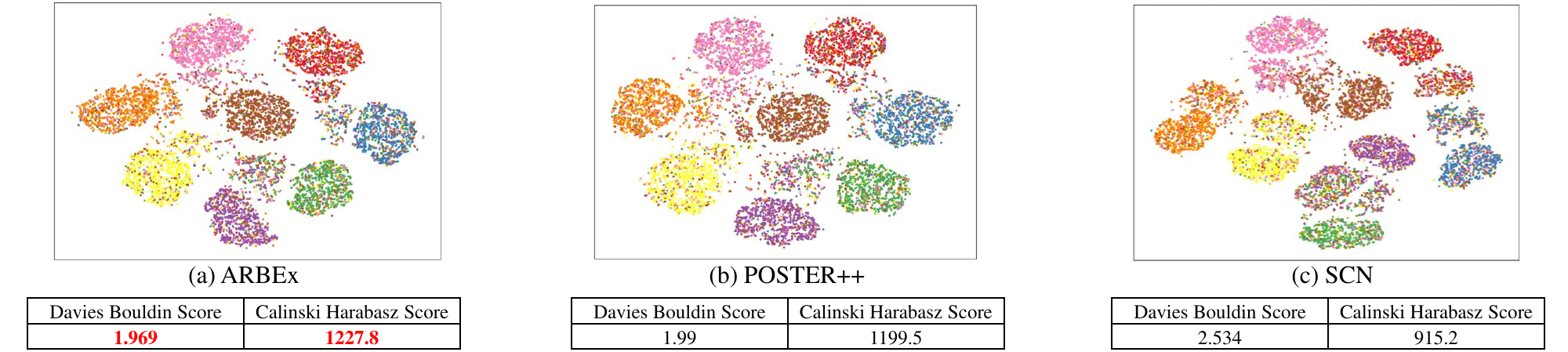}}
\caption{\textbf{t-SNE visualization} of Embeddings with \textbf{Davies Bouldin Score ($\downarrow$) and Calinski Harabasz Score ($\uparrow$)} of our model \textbf{ARBEx} comparing with POSTER++ and SCN using \textit{Aff-Wild2} dataset containing 8 classes.}\label{fig:tSNE}
\end{figure*}

\subsection{Comparison with State-of-the-Art Methods}
Using five separate datasets — AffWild2, RAF-DB, JAFFE, FERG-DB, and FER+ (explained in \ref{Datasets}), the table \label{tab:Comparison of Accuracy} shows the comparison of the accuracy of multiple State-of-the-Art facial expression learning methods. In this study, the models SCN \cite{wang2020suppressing}, RAN \cite{wang2020region}, RUL \cite{zhang2021relative}, EfficientFace \cite{zhao2021robust}, POSTER \cite{zheng2022poster}, POSTER++ \cite{mao2023poster}, and ARBEx are compared.

\begin{table}[h]
\centering
\caption{Comparison of Accuracy (\%) ($\uparrow$) with SOTAs.  [KEY: \textcolor{red}{\textbf{Best}}
*= Our evaluation of the proposed model]}
\label{tab:Comparison of Accuracy}
\begin{tabular}{l c c c c c}
\hline
& \multicolumn{5 }{c}{Datasets} \\ \hline
Models  & AffWild2 & RAF-DB & JAFFE & FER+ & FERG \\ \hline
SCN \cite{wang2020suppressing} & 60.55 & 87.03 & 86.33 & 85.97 & 90.46 \\ \hline
RAN \cite{wang2020region} & 59.81 & 86.9 & 88.67 & 83.63 & 90.22 \\ \hline
RUL \cite{zhang2021relative} & 62.37 & 88.98 & 92.33 & - & 92.35 \\ \hline
EfficientFace \cite{zhao2021robust} & 62.21 & 88.36 & 92.33 & - & 92.16 \\ \hline
POSTER \cite{zheng2022poster} & 68.83 & 92.05 & 96.67 & 91.62 & 96.87 \\ \hline
POSTER++ \cite{mao2023poster} & 69.18 & 92.21 & 96.67 & 92.28 & 96.36 \\ \hline
\textbf{ARBEx} (Our) & \textcolor{red}{\textbf{72.48}} & \textcolor{red}{\textbf{92.47}} & \textcolor{red}{\textbf{96.67}} & \textcolor{red}{\textbf{93.09}} & \textcolor{red}{\textbf{98.18}} \\ \hline
\end{tabular}
\end{table}

Upon investigation of the results, it is apparent that ARBEx outperforms all other models across all datasets, attaining the highest accuracy scores for each dataset. Specifically, ARBEx earns an accuracy score of 72.48\% on the AffWild2 \cite{kollias2023abaw1} dataset, which is significantly higher than POSTER++, which has an accuracy score of 69.18\%. ARBEx outperforms every other model in the study, with accuracy scores on the RAF-DB \cite{li2017reliable, li2019reliable}, FERG-DB \cite{aneja2016modeling}  and JAFFE \cite{lyons2020coding, lyons2021excavating} datasets of 92.47\%, 98.18\% and 96.67\%, respectively. Finally, on the FER+ \cite{BarsoumICMI2016} dataset, ARBEx acquires an accuracy score of 93.09\%, which significantly outperforms every other model tested. Our novel reliability balancing section reduces all kinds of biases, resulting in exceptional performance in all circumstances.

\subsection{Visualization Analysis}
\subsubsection{Confidence Probability Distributions}

Some working examples of reliability balancing method are added in Fig. \ref{fig:3}. It demonstrates how the method enhances the accuracy and reliability of predictions by stabilizing probability distributions. The primary probabilities generated by the model exhibit considerable variation, ranging from 0.529 to 0.0026. Due to intra-class similarity, disparity, and label ambiguity issues within images, primary probabilities are not always reliable in predicting accurate labels. In Fig. \ref{fig:3}, it is evident that the maximum primary probability exceeds 0.4 for most of the cases, despite the associated labels being erroneous and making the model unreliable. Upon implementing the correction method, we notice two different phenomena.

\textbf{Boost in confidence values of accurate labels.} We observe a rise in maximum confidence value in probability distribution in some cases (Label 2, 5, 7 in Fig. \ref{fig:3}) after applying reliability balancing. Increased confidence probability ensures accurate predictions.

\textbf{Decrease in confidence in faulty labels.} In some other cases, the confidence levels of faulty predictions are decreased by reliability balancing. In these situations, incorrect maximum values are reduced to a range of 0.15-0.25 mostly (Label 0, 1, 3 in Fig. \ref{fig:3}). The correct maximum values stay in a range of 0.2-0.3, while also providing the correct labels. These findings further support the vital role of reliability balancing and stabilization techniques. 

By implementing the corrective measures afforded by reliability balancing, both the maximum and minimum probability increases to 0.5429 and 0.0059 across a thorough study sample of 60 images, stabilizing the distribution. Notably, the standard deviation of corrected predictions (0.0881) was found to be lower than that of primary predictions (0.1316), providing strong evidence for enhanced stability and balance proving the efficacy of reliability balancing method applied. Hence, the reliability balancing strategy supports the model in all circumstances, from extremely uncertain conditions to extremely confident scenarios, whenever the primary model is making poor conclusions.

\subsubsection{Clustering Embeddings}
The t-distributed stochastic neighbor embedding (t-SNE) plot in Fig. \ref{fig:tSNE} visualizes the difference between classes in embedding space, each color denoting one class. \textbf{Davies-Bouldin score}\cite{davies_bouldin_score1979} gauges the mean resemblance between a given cluster and its most akin cluster. Lower scores implies better clustering outcome. \textbf{Calinski-Harabas score}\cite{calinski_harabasz_score1974} assesses the ratio of variance between clusters to variance within clusters. A higher score suggests an optimal clustering solution. 

In the figures, we observe some uniformly spaced groups with reliable classifications and some noises denoting problematic circumstances with inter-class similarity and disparity issues. The scores indicate that ARBEx outperforms the other models in terms of both Davies Bouldin Score (1.969) and Calinski Harabasz Score (1227.8). These results indicate that ARBEx produces a better clustering outcome compared to POSTER++ (Davies Bouldin Score of 1.990 and Calinski Harabasz Score of 1199.5) and SCN (2.534 and 915.2). Based on the plots and scores, it is noticeable that ARBEx has embeddings that are well dispersed and more discriminating than POSTER++ and SCN.

\begin{figure}[h!] 
\centering{\includegraphics[scale=0.37]{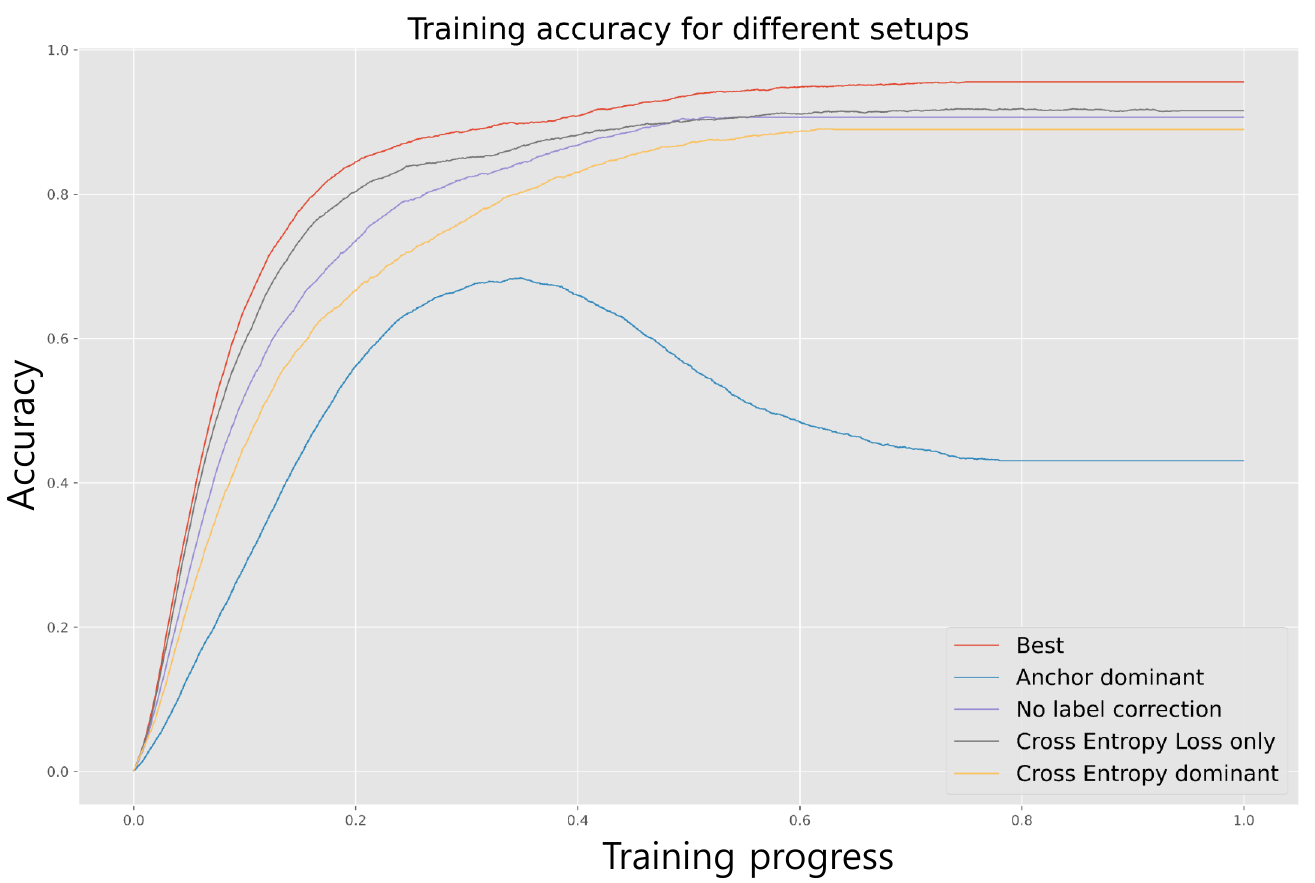}}
\caption{Study of training progress on different setups using Accuracy (\%) score. The red line shows the optimal model with perfect loss combination, blue line shows anchor loss dominant model, indigo colored line shows the model with no label correction with anchors, the grey line shows the model with Cross-Entropy Loss only and the yellow line shows where Cross-Entropy Loss is dominant.}\label{fig:5}
\end{figure}

\subsubsection{Study of Different Loss Functions}
Fig. \ref{fig:5} demonstrates the effects of different loss function setups in the training stage of our experiment. Anchor loss dominance causes the model to drop its performance after some initial good epochs, conveying that the model starts overfitting on anchors, ignoring true labels.  Relying more on similarities rather than the actual prediction performance, this setup fails to fulfill the criteria. The other setups are quite stable and close. The ideal combination used in the study helps the model to train faster and better.

\section{Conclusion}

In this paper, we have presented a novel approach \textbf{ARBEx} for FEL, which leverages an extensive attentive feature extraction framework with reliability balancing to mitigate issues arising from biased and unbalanced data. Our method combines heavy augmentation and data refinement processes with a cross-attention window-based Vision Transformer (ViT) to generate feature embeddings, enabling effective handling of inter-class similarity, intra-class disparity, and label ambiguity. Our unique reliability balancing strategy combines trainable anchor points in the embedding space and multi-head self-attention mechanism with label distributions and confidence to stabilize the distributions and maximize performance against poor projections. Experimental analysis across multiple datasets demonstrates the superior effectiveness of our proposed ARBEx method, outperforming state-of-the-art FEL models, thereby highlighting its potential to significantly advance the field of facial expression learning.

\ifCLASSOPTIONcompsoc
  \section*{Acknowledgments}
\else
  \section*{Acknowledgment}
\fi
This work was partly supported by (1) Institute of Information \& communications Technology Planning \& Evaluation (IITP) grant funded by the Korea government (MSIT) (No.2020-0-01373, Artificial Intelligence Graduate School Program (Hanyang University)) and (2) the Bio \& Medical Technology Development Program of the National Research Foundation (NRF) funded by the Korean government (MSIT) (No. NRF-2021M3E5D2A01021156).

{\small
\bibliographystyle{ieee_fullname}
\bibliography{main}
}

\end{document}